\documentclass[runningheads]{llncs}
\usepackage[T1]{fontenc}
\usepackage{amsmath}
\usepackage{amsfonts}

\usepackage{graphicx,verbatim}
\usepackage{multirow}
\begin{document}
\title{USEMA: a Scalable Efficient Mamba Like Attention for Medical Image Segmentation}
%\titlerunning{Abbreviated paper title}
% If the paper title is too long for the running head, you can set
% an abbreviated paper title here
%
\begin{comment}  %% Removed for anonymized MICCAI submission
\author{First Author\inst{1}\orcidID{0000-1111-2222-3333} \and
Second Author\inst{2,3}\orcidID{1111-2222-3333-4444} \and
Third Author\inst{3}\orcidID{2222--3333-4444-5555}}
%
\authorrunning{F. Author et al.}
% First names are abbreviated in the running head.
% If there are more than two authors, 'et al.' is used.
%
\institute{Princeton University, Princeton NJ 08544, USA \and
Springer Heidelberg, Tiergartenstr. 17, 69121 Heidelberg, Germany
\email{lncs@springer.com}\\
\url{http://www.springer.com/gp/computer-science/lncs} \and
ABC Institute, Rupert-Karls-University Heidelberg, Heidelberg, Germany\\
\email{\{abc,lncs\}@uni-heidelberg.de}}

\end{comment}
\author{Elisha Dayag \inst{1} \and
Nhat Thanh Tran \inst{1} \and
Jack Xin \inst{1}}
\institute{University of California Irvine, Irvine CA 92697}
\authorrunning{E. Dayag et al.}
  %% Added for anonymized MICCAI submission

\maketitle              % typeset the header of the contribution
%
%While vision transformers are able to capture global interactions using vanilla self-attention, their quadratic computational complexity in the input size remains a struggle for medical image segmentation tasks. Linear attention variants offer an alternative linear complexity however they struggle with recall, hindering their effectiveness in computer vision tasks
\begin{abstract}
Accurate medical image segmentation is an integral part of the medical image analysis pipeline that requires the ability to merge local and global information. While vision transformers are able to capture global interactions using vanilla self-attention, their quadratic computational complexity in the input size remains a struggle for medical image segmentation tasks. Motivated by the dispersion property of vanilla self-attention and recent development of Mamba form of attention, Scalable and Efficient Mamba like Attention (SEMA) utilizes token localization via local window attention to
avoid dispersion and maintain focusing, complemented by theoretically consistent
arithmetic averaging to capture global aspect of attention. In this work, we present USEMA, a hybrid UNet architecture that merges the local feature extraction ability of convolutional neural networks (CNNs) with SEMA attention. We conduct experiments with USEMA across a variety of modalities and image sizes, demonstrating improved computational efficiency compared to transformer based models using full self-attention, and superior segmentation performance relative to purely convolution and Mamba-based models. 

% Authors must provide keywords and are not allowed to remove this Keyword section.

\end{abstract}
\section{Introduction}
Medical image segmentation is one of the fundamental steps of the medical image analysis pipeline. Accurate segmentation results can assist doctors in providing diagnoses \cite{rayed2024deep}. Medical images present unique challenges in segmentation due to the limited amount of data available for training as well as the complex, overlapping geometry of the objects to be segmented \cite{ronneberger2015u}.

Inspired by the state of the art performance of Transformer-based frameworks on various computer vision tasks \cite{dosovitskiy2020image}, myriad works have studied transformers for medical image segmentation. Many of these works combine utilize hybrid transformer/CNN architectures to utilize the latter's ability to efficiently extract image features \cite{hatamizadeh2021swin,hatamizadeh2022unetr,zhou2023nnformer}. While vanilla self-attention has a global receptive field and enables a model to capture long-range dependencies, it incurs an $O(n^2)$ sequence complexity ($n$ the sequence length) that can be computationally prohibitive. Thus, many works have sought to provide the benefits of attention in a more efficient manner \cite{katharopoulos2020transformers,wang2020linformer,zhou2021informer}. 

A recent line of research utilizes state space models like Mamba \cite{gu2024mamba}. These models achieve the global receptive field and dynamic weighting parameters of self-attention with linear time complexity. As of now, multiple papers have been written incorporating Mamba for medical image analysis and demonstrating a marked improvement over the aforementioned attention-based models \cite{ma2024u,wang2024mamba,xing2025segmamba}. At the same time, several studies have utilized aspects of the Mamba architecture to derive efficient attention variants \cite{han2024demystify,li2025transmamba,tran2025semascalableefficientmamba}. Some of these Mamba-inspired variants have been utilized in medical image segmentation to great success \cite{jiang2024mlla}.  

In this paper we propose a \textbf{UNet} with \textbf{S}calable \textbf{E}fficient \textbf{M}amba-Like \textbf{Attention} (USEMA). Our model builds upon the SEMA mechanism \cite{tran2025semascalableefficientmamba} and utilizes a local-global attention that carefully augments window attention with a theoretically-grounded efficient global attention approximation. By combining this with the adaptive feature selection capabilities of Mamba and integrating the mechanism into a symmetric U-shaped structure, we obtain an architecture that effectively utilizes SEMA's strengths for the particular task of medical image segmentation. Our results show that USEMA  outperforms a number of both Transformer-based and Mamba-based medical image segmentation models across a diverse selection of 2D benchmarks. 
\section{Methods}

\subsection{Preliminary}

Transformer \cite{vaswani2017attention} has been a dominant architecture in deep learning. The main component is attention mechanism. For given input $x \in\mathbb{R}^{n \times d}$, attention is defined as:
\begin{equation}
  A(Q, K , V) = \text{softmax}(QK^T)V, 
\end{equation}
where $Q = xW_Q + b_Q, K = xW_K + b_K, V = xW_V + b_V$. Here $W_Q, W_K, W_V \in\mathbb{R}^{d\times d}$ and $b_Q, b_K, b_V \in\mathbb{R}^{n\times d}$ are learnable parameters. An advantage of the attention mechanism is the ability to select relevant information from the given global context. However, there are two large problems with the use of attention in medical image segmentation. The first and most often referenced is the quadratic computational cost of attention with respect to sequence length \cite{gu2024mamba,hatamizadeh2021swin,liu2024swin,wang2024mamba,xing2025segmamba}. Another issue is that as the sequence length increases, the scores of the attention matrix (softmax($(QK^T)$)) disperse toward 0 uniformly, which hinders the attention's ability to select important keys and values \cite{tran2025semascalableefficientmamba}. To be more precise, under suitable conditions \cite{tran2025semascalableefficientmamba} proves that there exists an $N_0 \in\mathbb{N}$ such that for all $n > N_0$ we have 
\begin{equation}
\dfrac{C_1}{n}< \text{softmax}((QK^T))_{ij} < \dfrac{C_2}{n},    \label{disp_ineq}
\end{equation}
for all indices $0\le i,j \le n$, and for some $C_1, C_2 \in \mathbb{R}^{+}$. This holds for all attention heads in a transformer. A related probabilistic result also showed \cite{tran2025semascalableefficientmamba} that if entries of the attention matrix are random variables 
%(not necessarily i.i.d) and the scores are 
allowed to be mildly correlated, dispersion occurs with high probability as a consequence of the weak law of large numbers.

\begin{figure}
    \centering
    \includegraphics[width=.9\linewidth]{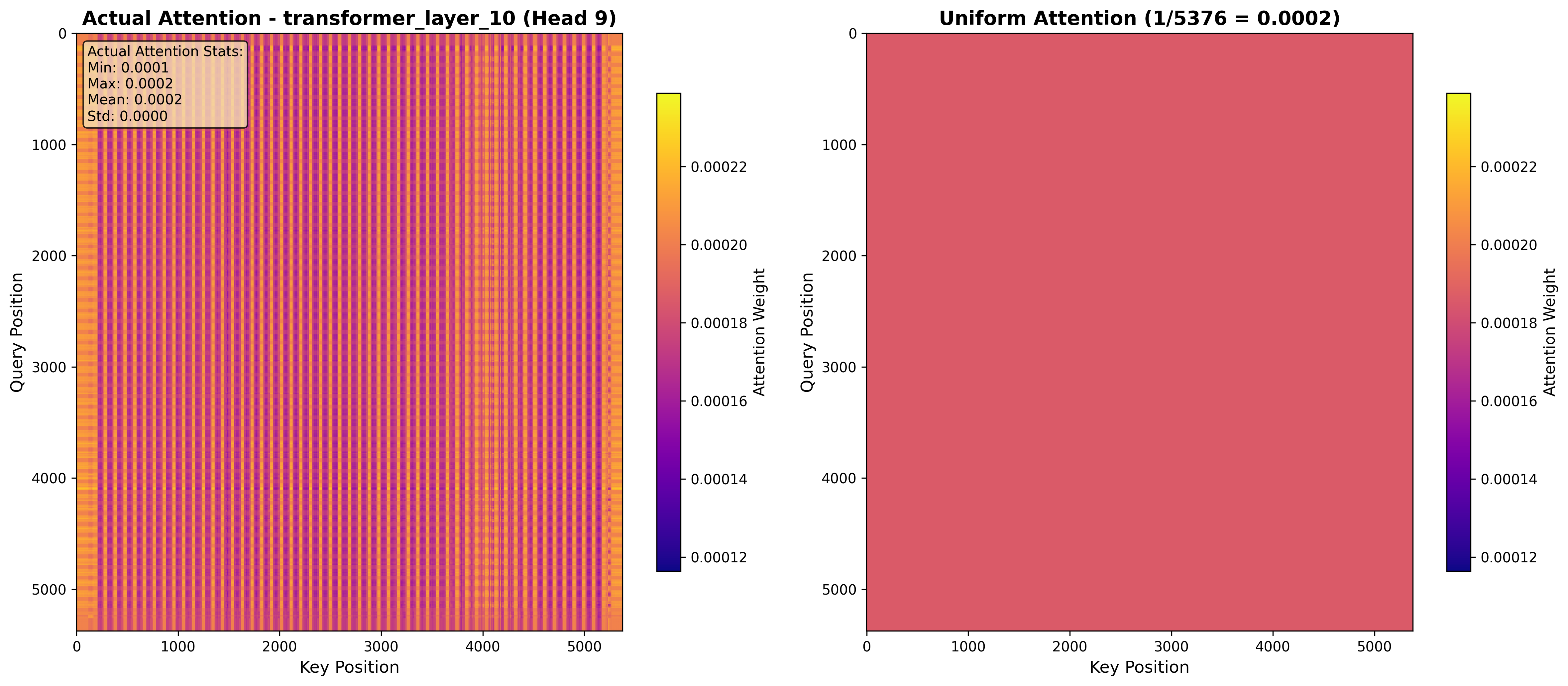}
    \caption{Left: The attention matrix of UNETR on a sequence of length 5376 Right: The attention matrix that would be obtained via a uniform attention i.e. setting each entry to $1/5376$. Note that the values in the left-hand matrix are all within $1e-4$ of the mean and bounded above by $2e-4$ and below by $1e-4$.}
    \label{unetr dispersion}
\end{figure}

We can observe experimentally the dispersion phenomenon described theoretically above. We trained UNETR \cite{hatamizadeh2022unetr}, which utilizes full self-attention, on images from the MICCAI 2017 Endovis Challenge \cite{allan20192017}. We trained on image patches of size $1536 \times 896$, which with the standard $16\times 16$ patches of a vision transformer leads to a sequence of length $96\times 56 = 5376$. When we visualize the attention matrices produced by UNETR on test data, as seen in Figure \ref{unetr dispersion}, we see that the attention scores are remarkably concentrated around the mean, $\frac{1}{\text{seq len}}$. 

A way to mitigate the computational cost is to limit the context window of the selection process, that is window attention defined as:
%\begin{equation}\label{eq:win_attn_proof}
%        A_w(Q, K, V):= 
%        \begin{bmatrix}
%              \dfrac{\sum_{j\in J(1)}\exp(q_1k_j^T)v_j}{\sum_{i\in J(1)} \exp(q_1k_i^T)}\\
 %           \vdots \\
 %            \dfrac{\sum_{j\in J(n)}\exp(q_nk_j^T)v_j}{\sum_{i\in J(n)} \exp(q_nk_i^T)}
  %      \end{bmatrix},
  %  \end{equation}
\begin{equation}\label{eq:win_attn_proof}
        A_w(Q, K, V):= 
        \begin{bmatrix}
              \dfrac{\sum_{j\in J(1)}\exp(q_1k_j^T)v_j}{\sum_{i\in J(1)} \exp(q_1k_i^T)},
            \cdots,
             \dfrac{\sum_{j\in J(n)}\exp(q_nk_j^T)v_j}{\sum_{i\in J(n)} \exp(q_nk_i^T)}
        \end{bmatrix}^{T},
    \end{equation}
for some index set $J(m)$. An example is $J(m) = \{Mw +1,\dots, (M+1)w\}$, where $M = \lfloor\frac{m-1}{w}\rfloor$. This reduces the computational cost to linear with respect to the input sequence length. 

As an alternative to window attention, state space models, in particular Mamba, have been a competitive replacement with its linear computational cost \cite{gu2024mamba,NEURIPS2020_ssm_hippo}. Concretely, Mamba is defined as a map from $x$ to $y$ given by the following dynamical equations:
\begin{equation}\label{eq:mamba}
h_t = A_t \odot  h_{t-1} + B_t( \Delta_t \odot x_t), y_t = C_t h_t + D \odot x_t,
\end{equation}
where $x_t, \Delta_t \in \mathbb{R}^{1\times d}, A_t, h_t \in \mathbb{R}^{d\times d}, B_t \in\mathbb{R}^{d\times 1}$ and $y_t\in \mathbb{R}^{1\times d}, C_t \in\mathbb{R}^{1\times d}, D\in\mathbb{R}^{1\times d}$, and $\odot$ denotes the Hadamard product. By induction one can show that 
% \begin{equation}\label{eq:mamba_linear_attn_rep}
    $y_m = \sum_{i=1}^m q_m \Tilde{k}_i^T\Tilde{v}_i + D\odot v_m,$
% \end{equation}
where $\Tilde{k}_i^T \Tilde{v}_i= \left(\prod_{j=1}^{m-i} A_{m-(j-1)}\right) \odot (B_i(\Delta_i \odot x_i))$, $q_m \simeq C_m$, and $v_m \simeq x_m$ \cite{tran2025semascalableefficientmamba}. The former can be understood as causal un-normalized masked attention, and the latter is the skip connection. We observe that in practice $A_m$ entries range from $0$ to $1$, then the associated casual un-normalized attention displays exponential forgetting with respect to $m$.
 
Inspired by Mamba's exponential forgetting which mimics window attention, and the almost uniform attention score distribution due to long sequence, SEMA adopted window attention as a local approximator and the average as the global approximation to softmax attention, in particular 
\begin{equation}\label{SEMA}
    SEMA(Q,K,V) := A_w(Q,K,V) + \left[\frac{1}{n} \sum_{j=1}^n v_j \right],
\end{equation}
where $\left[ \cdot \right]$ broadcasts the row $n$ times. The global arithmetic average term 
in (\ref{SEMA}) reflects the $O(1/n)$ bounds in the dispersion inequality 
(\ref{disp_ineq}).
SEMA is then inserted into a Mamba macro structure (see SEMA block, Fig. 2). We will demonstrate that the average contributes to a very cost effective global approximation for the full attention in Section \ref{sec:ablation}.

\begin{figure}[ht]
    \centering
    \begin{minipage}{0.45\textwidth}
        \centering
\includegraphics[width=\linewidth]{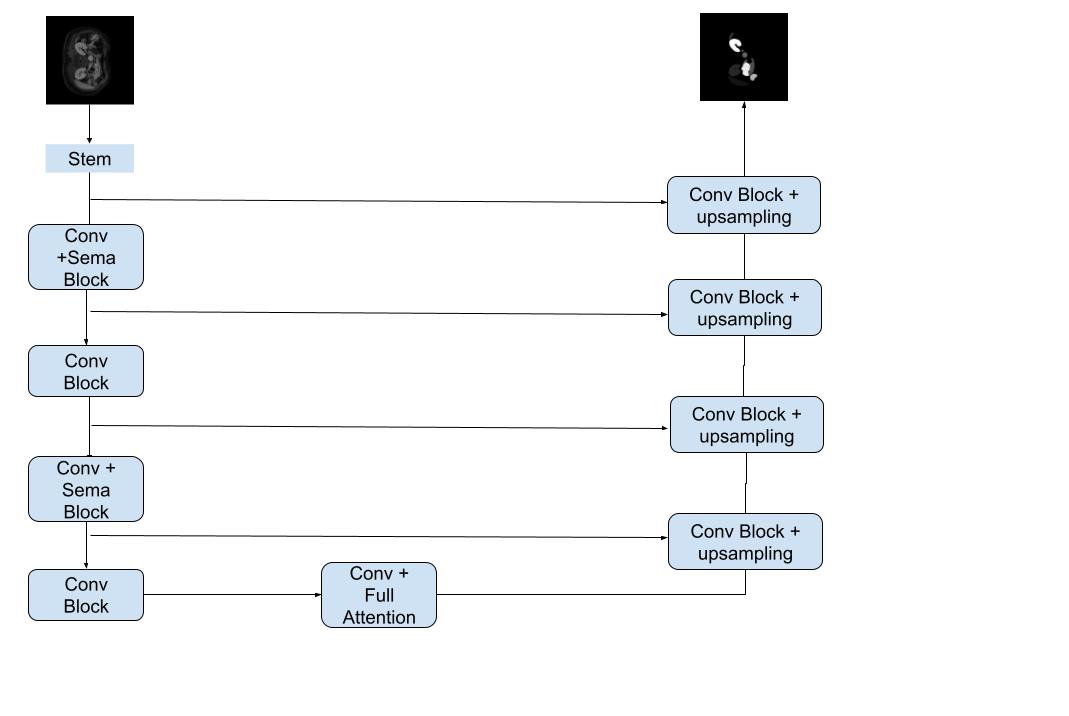}
    \end{minipage} 
    \hfill
    \begin{minipage}{0.45\textwidth}
        \centering
        \includegraphics[width=\linewidth]{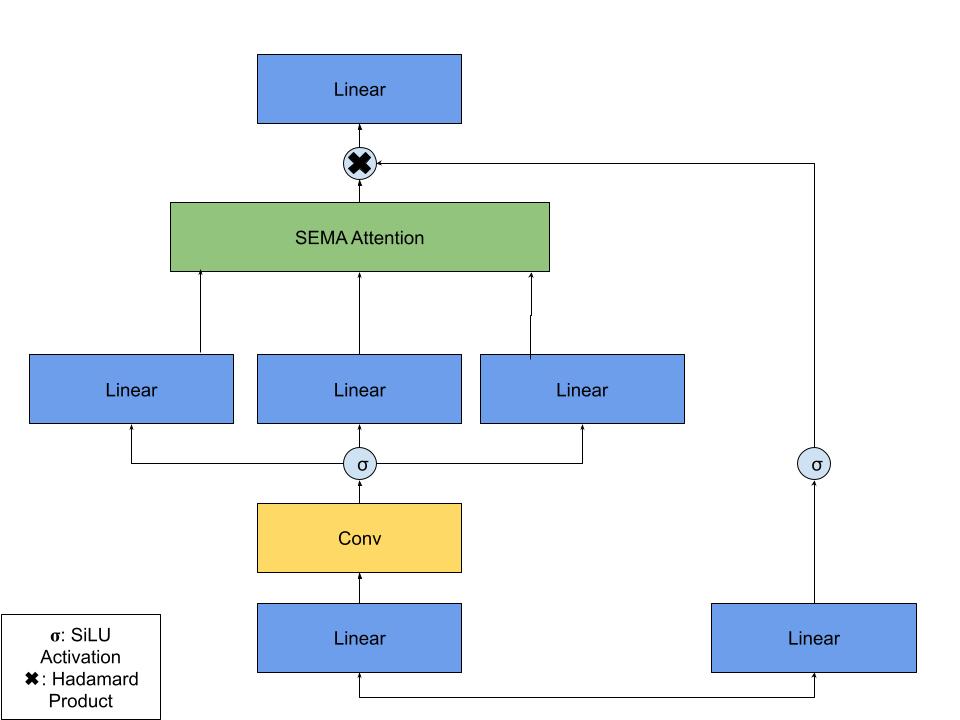}
    \end{minipage}
    \caption{USEMA architecture (left) and its SEMA block (right).}
    \label{sema_figure}
\end{figure}
\subsection{USEMA: Merging SEMA with UNet}To utilize SEMA for medical image segmentation, we merge SEMA into the UNet framework to create USEMA, depicted in Fig. \ref{sema_figure}. Each building block contains two residual blocks followed by a SEMA block. The residual block contains a plain convolutional layer followed by instance normalization \cite{ulyanov2016instance} and Leaky ReLU \cite{maas2013rectifier}. The image features are then reshaped from $(B,C,H,W)$, where B is the batch size and C is the number of channels, to $(B,C,HW)$ before being passed to the SEMA block. 

In the SEMA block, the features  first pass through conditional positional embedding \cite{chu2021conditional} before undergoing layer normalization \cite{ba2016layer}. Next the features pass through two branches: in one branch a linear layer followed by SiLU is applied to serve as a gating mechanism similar to Mamba.  In the other branch, a linear layer followed by depthwise convolution is applied to the features, followed by the SEMA attention in Eq. (\ref{SEMA}). Aside from the local-global attention, the features are also enhanced with Rotary Positional Embedding \cite{su2024roformer} and locally-enhanced position encodings \cite{dong2022cswin}. After undergoing SEMA attention, the two branches are merged via the Hadamard product before being enhanced with another conditional positional encoding to inject spatial position awareness into the post-attention features. Finally the features pass through a feedforward network before moving on to the next block.  

In the bottleneck, the spatial dimensions of the image features have been compressed enough so that it is unnecessary computationally to utilize windowed attention. Thus in the bottleneck we use full self-attention. 

The decoder consists only of residual blocks and transposed convolutions. As is typical with the UNet structure, we utilize skip connections to connect hierarchical features between the encoder and decoder via concatenation. To produce the segmentation output, we apply a $1\times 1$ convolution to the final decoder feature. 
%\vspace{-0.1 in}

\section{Experiments}
%\vspace{-0.1 in}

\subsection{Datasets}
To validate the performance of USEMA, we benchmarked it on three datasets spanning multiple imaging modalities, resolutions, and segmentation targets. All of our data preprocessing and loading was done with the nnUNet framework to ensure fair comparisons between our model and benchmarks.  

\textbf{Abdomen MRI:} This dataset was originally from the MICCAI 2022 AMOS Challenge\cite{ma2024unleashing}. It focuses on segmenting 13 abdominal organs from MRI scans, including the liver, right kidney, left kidney, spleen, pancreas, aorta, inferior vena cava, right adrenal gland, left adrenal gland, gallbladder, esophogous, stomach, and duodenum. We followed the settings in \cite{ma2024u} and used the original 60 images containing 5615 slices for training and the additional 50 scans containing 3357 slices for testing. We cropped the images into patches of size $(320,320)$ for training and testing.

\textbf{Endoscopy images:} This dataset was originally from the MICCAI 2017 Endovis Challenge \cite{allan20192017}. The training and testing data contained image frames from endoscopy videos and the task focused on segmenting seven different instrument types, namely the large needle driver, prograsp forceps, monopolar curved scissors, cadiere forceps, bipolar forceps, vessel sealer, and drop-in ultrasound probe. We followed the official dataset splitting of 1800 images in the training set and 1200 in the test set. In line with other works \cite{liu2024swin,ma2024u}, images were cropped to $(384,640)$. Notably, this dataset contains a unique aspect ratio compared to the other datasets. 

\textbf{Microscopy images:} This dataset is from the NeurIPS 2022 Cell Segmentation Challenge \cite{ma2024multimodality}, which focused on cell segmentation in microscopy images. It consists of 1000 images for training and 101 images for testing. Unlike the other two datasets, this task is an instance segmentation task. The images in this dataset were cropped to $(512,512)$.
%\vspace{-0.1 in}

\subsection{Implementation Details}
To ensure consistency across all  datasets, we strictly adhere to the nnUNet-framework \cite{isensee2021nnu} for preprocessing. For the microscopy and endoscopy datasets, we used a window size of 16 in our sema block. On the Abdomen MR dataset we used a window size of 32. All other model hyperparameters remained consistent across all datasets. To train our model, we used the AdamW optimizer with learning rate $3e-4$ and weight decay $0.05$. The loss function is the sum of dice loss and cross-entropy loss and we perform deep supervision \cite{pmlr-v38-lee15a} at each scale.  For each dataset we trained for 1000 epochs using a cosine annealing learning rate scheduler with $T_{\max} = 100$. 
%\vspace{-0.1 in}

\subsection{Benchmarks and Evaluation Details}
To assess the performance of our model, we compare it to a variety of transformer-based methods and Mamba-based or Mamba-inspired methods. The transformer-based methods compared against are UNETR, \cite{hatamizadeh2022unetr}, Swin-UNETR \cite{hatamizadeh2021swin}, and nnFormer \cite{zhou2023nnformer}. The Mamba-based methods we compare against are U-Mamba \cite{ma2024u}, Swin-UMamba\cite{liu2024swin}, and Mamba UNet \cite{wang2024mamba}. For U-Mamba, we compare against the variant U-Mamba\_Enc as it is most similar in structure to our USEMA. Similarly for Swin-UMamba, we only compare to the version utilizing the 
Swin-Umamba encoder. 

Dice similarity coefficient (DSC) and normalized surface distance (NSD) were used to evaluate performance on the abdomenMRI and Endoscopy datasets. The microscopy dataset is an instance segmentation task, so we use F1 score for evaluation. We also compute the number of parameters ($\#$ param) of each model. Baseline results for DSC, NSD, and F1 score for a number of our models are referenced from \cite{liu2024swin,zhang2025switch} except for MLLA-UNet, whose results we report based on their official implementation.
%\vspace{-0.1 in}

\subsection{Comparisons on AbdomenMRI Dataset} 
Table \ref{tab:Abdomen} displays the segmentation results on the Abdomen dataset. USEMA outperforms all the comparative baselines, comprising both Mamba-based/inspired and transformer-based methods. In particular, when compared to the other transformer-based methods, our model enhances the DSC and NSD by $5.83\%$ and $4.79\%$ respectively. Additionally, our model is $13\%$ smaller than the best performing transformer-based model. Our model also outperforms all the Mamba-based methods, enhancing the DSC and NSD by $10.53\%$ and $9.60\%$ respectively over the other Mamba-inspired attention variant in MLLA-UNet.
    \begin{table}[h]
        \centering
        \scalebox{0.95}{\begin{tabular}{|c|c|c|c|}
        \hline
             Model &  DSC $(\uparrow) $& NSD $(\uparrow)$ & \# Params \\
             \hline
            USEMA (ours) & {\bf .7704} & {\bf .8345} & 52M\\
            \hline
            U-Mamba Enc \cite{ma2024u}& .7625 & .8327 & 67M \\ 
            \hline
            Mamba UNet \cite{wang2024mamba}& .7496 & .8178 & 35M \\
            \hline
            Swin U-Mamba \cite{liu2024swin}& .7054 & .7647 & 60M \\ 
            \hline
            MLLA-UNet \cite{jiang2024mlla}& .6970 & .7614 & 47M\\
            \hline
            UNETR \cite{hatamizadeh2022unetr}& .5747 & .6309 &  87M\\ 
            \hline
            SwinUNETR \cite{hatamizadeh2021swin}& .7028 & .7669 & 25M\\
            \hline
            nnFormer \cite{zhou2023nnformer}& .7279 & .7963 & 60M\\ 
            \hline
        \end{tabular}}
        
        \caption{Results of organ segmentation on the MICCAI 2022 Abdomen MRI dataset. For a fair
comparison, the Swin-UMamba results are reported without the benefit of
ImageNet-based pre-training }
        \label{tab:Abdomen}
    \end{table}
%\vspace{-0.5 in}

\subsection{Comparisons on Endoscopy Dataset}
Table \ref{tab:endo} shows the segmentation results on the endoscopy dataset. We improve upon the transformer-based methods by $5.63\%$ and $6.31\%$ in DSC and NSD respectively. Our model also outperforms the Mamba-based methods, in particular enhancing the DSC and NSD by $13.49\%$ and $13.51\%$ respectively over the other Mamba-inspired attention variant in MLLA-UNet. 
    \begin{table}[h]
        \centering
        \scalebox{0.95}{\begin{tabular}{|c|c|c|c|}
        \hline
             Model &  DSC $(\uparrow) $& NSD $(\uparrow)$ & \# Params \\
             \hline
            USEMA (ours) & {\bf .6463} & {\bf .6621} & 52M\\
            \hline
            U-Mamba Enc \cite{ma2024u}& .6303 & .6451 & 67M\\ 
            \hline
            Mamba UNet \cite{wang2024mamba}& .6256 & .6370 & 35M\\
            \hline
            Swin U-Mamba \cite{liu2024swin}& .6402 & .6547 & 28M\\ 
            \hline
            MLLA-UNet \cite{jiang2024mlla}& .5695 &.5833 & 47M\\
            \hline
            UNETR \cite{hatamizadeh2022unetr}& .5017 & .5168 & 88M\\ 
            \hline
            SwinUNETR \cite{hatamizadeh2021swin}& .5528 & .5683 & 25M\\
            \hline
            nnFormer \cite{zhou2023nnformer}& .6135 & .6228 & 60M\\ 
            \hline

        \end{tabular}}
        
        \caption{Results of instruments segmentation on the Endoscopy 2017 dataset. For a fair
comparison, the Swin-UMamba results are reported without the benefit of
ImageNet-based pre-training }
        \label{tab:endo}
    \end{table}
\vspace{-0.3 in}

\subsection{Comparisons on Microscopy Dataset}
Table \ref{tab:cell} displays the segmentation results on the microscopy dataset. Compared to the other datasets, this dataset has fewer images, higher image resolution, and larger visual differences. These constraints require the model to capture long-range dependencies and learn in a data efficient manner. We can see that our model outperforms all of the baseline competitors. Among transformer-based models, USEMA enhances the F1 score by $8.61\%$. USEMA surpasses the other Mamba-inspired architecture MLLA-UNET, by $19.23\%$ in F1 score. Additionally, USEMA manages to have $43\%$ fewer parameters than U-Mamba\_Enc, the strongest Mamba-based model we compared against.   

    \begin{table}[h]
        \centering
        \scalebox{0.95}{\begin{tabular}{|c|c|c|}
        \hline
             Model &  F1 $(\uparrow)$ & \# Params \\
             \hline
            USEMA (ours) & {\bf .5791} & 52M\\
            \hline
            U-Mamba Enc \cite{ma2024u} & .5607 & 92M\\ 
            \hline
            Mamba UNet \cite{wang2024mamba}& .5215 & 35M\\
            \hline
            Swin-UMamba \cite{liu2024swin}& .5186 & 60M\\ 
            \hline
            MLLA-UNet \cite{jiang2024mlla}& .4857 & 47M \\
            \hline
            UNETR \cite{hatamizadeh2022unetr}& .4357  & 88M\\ 
            \hline
             SwinUNETR\cite{hatamizadeh2021swin}& .3967 & 25M\\
            \hline
            nnFormer \cite{zhou2023nnformer} & .5332 & 60M\\ 
            \hline
        \end{tabular}}
        
        \caption{Results of cell segmentation on the Microscopy dataset. For a fair comparison,
the Swin-UMamba results are reported without the benefit of ImageNet-based
pre-training. }
        \label{tab:cell}
    \end{table}
%\vspace{-0.1 in}
    
\section{Ablation Study}\label{sec:ablation}
In this section, we analyze the impact of the global attention approximation in Eq. \ref{SEMA}. We retrain USEMA on the same datasets using only the window attention. The results are reported in table \ref{tab:ablation}. We can see that the inclusion of the global attention approximation enhances our models segmentation ability.

\begin{table}[h]
    \centering
    \scalebox{0.95}{\begin{tabular}{|c|c|c|c|}
    \hline
        Dataset & Model & DSC/F1 $(\uparrow)$ & NSD $(\uparrow)$ \\
        \hline \multirow{2}{6em}{Abdomen MRI} & USEMA  & .7704 & .8345 \\
         & USEMA w/o attention approx &.7574 & .8214\\
         \hline \multirow{2}{6em}{Endoscopy} & USEMA  & .6463 & .6621 \\
         & USEMA w/o attention approx & .6218 & .6367\\
        \hline \multirow{2}{6em}{Microscopy} & USEMA  & .5791 & $-$\\
         & USEMA w/o attention approx & .5443 & $-$\\
         \hline
    \end{tabular}}
    \caption{Ablation study of homogeneous mixing (averaging) on the three datasets covered in this paper. In line with the results of Tabs. 1-3 above, we only present F1 score for the microscopy dataset. }
    \label{tab:ablation}
\end{table}
%\vspace{-0.1 in}

\section{Conclusions}
We embedded a mathematically guided scalable and efficient Mamba-like attention (SEMA) in a hybrid Unet architecture to integrate local feature extraction and global information coupling for medical image segmentation. The resulting network USEMA consistently out-performs  the state-of-the-art convolution and Mamba (linear time complexity recurrent attention \cite{gu2024mamba}) based models on benchmark admomen MRI, endoscopy and microscopy images, while saving computational costs compared to transformer based Unet with full attention. In future work, we plan to extend the arithmetic averaging in SEMA to learned and weighted averaging as well as sparsely weighted averaging in case of extremely long tokens, and  apply USEMA to large size medical images in pathology \cite{whole_slide_2024}.

\section{Acknowledgements}
The work was partially supported by NSF grants DMS2219904, DMS-2309520, and a Qualcomm Gift Award. NTT was also
funded by a Faculty Endowed Fellowship and the Graduate Scholar Success
Fund from UC Irvine.

%\begin{figure}
%\includegraphics[width=\textwidth]{fig1.eps}
%\caption{A figure caption is always placed below the illustration.
%Please note that short captions are centered, while long ones are
%justified by the macro package automatically.} %\label{fig1}
%\end{figure}

 %% removed for anonymized MICCAI submission.
    
    % The following acknowledgement and disclaimer sections can be removed for the double-blind review process.  If and when your paper is accepted, reinsert the acknowledgement and the disclaimer clause in your final camera-ready version.
    % IF you opted to include the acknowledgement and disclaimer sections, they will count towards the 8-page limit.

% ---- Bibliography ----
%
% BibTeX users should specify bibliography style 'splncs04'.
% References will then be sorted and formatted in the correct style.
%
%\newpage

 \bibliographystyle{splncs04}
 \bibliography{bibby}

\end{document}